\documentclass[journal ]{IEEEtran}
\usepackage{changepage}
\usepackage[normalem]{ulem}
\usepackage{graphicx}
\usepackage{epstopdf}
\usepackage{subfigure}
\usepackage{epsfig}
\usepackage{float}
\usepackage{multirow}
\usepackage{tabularx}
\usepackage{makecell}
\usepackage{booktabs}
\usepackage{dsfont}
\usepackage{color,hyperref}
\usepackage{amsmath}
\usepackage{amsthm}
\usepackage{amssymb}
\usepackage{threeparttable}
\usepackage[all]{xy}
\usepackage[lined,boxed,commentsnumbered, ruled]{algorithm2e}
\usepackage[dvipsnames]{xcolor}
\graphicspath{{./figs/}}
\newcommand{\citep}{\cite}
\theoremstyle{definition}
\newtheorem{defn}{Definition}
\newcommand{\argmin}{\operatornamewithlimits{argmin}}

\begin{document}
\title{Low-rank Multi-view Clustering in Third-Order Tensor Space}
\author{Ming Yin, \textit{Member, IEEE}, ~Junbin Gao,~Shengli Xie, \textit{Senior Member, IEEE}, and~Yi Guo, \textit{Member, IEEE} \IEEEcompsocitemizethanks{\IEEEcompsocthanksitem  Ming Yin and Shengli Xie are with School of Automation, Guangdong University of Technology, Guangzhou 510006, China. E-mail: \{yiming, shlxie\}@gdut.edu.cn  ~\textit{(Shengli Xie is the corresponding author of the paper.)}\protect \IEEEcompsocthanksitem Junbin Gao is with The Discipline of Business Analytics, The University of Sydney Business School, The University of Sydney, NSW 2006, Australia. E-mail: junbin.gao@sydney.edu.au. \protect \IEEEcompsocthanksitem Yi Guo is with School of Computing, Engineering and Mathematics, Western Sydney University, Parramatta, NSW 2150, Australia. E-mail:y.guo@westernsydney.edu.au}\thanks{Manuscript received xxxx; revised xxxx.}}

\IEEEcompsoctitleabstractindextext{%
\begin{abstract}
The plenty information from multiple views data as well as the complementary information among different views are usually beneficial to various tasks, e.g., clustering, classification, de-noising. Multi-view subspace clustering is based on the fact that
the multi-view data are generated from a latent subspace. To recover the underlying subspace structure, the success of the sparse and/or low-rank subspace clustering has been witnessed recently. Despite some state-of-the-art subspace clustering approaches can numerically handle multi-view data, by simultaneously exploring all possible pairwise correlation within views, the high order statistics is often disregarded which can only be captured by simultaneously utilizing all views. As a consequence, the clustering performance for multi-view data is compromised. To address this issue, in this paper, a novel multi-view clustering method is proposed by using \textit{t-product} in third-order tensor space. Based on the circular convolution operation, multi-view data can be effectively represented by a \textit{t-linear} combination with sparse and low-rank penalty using ``self-expressiveness''. Our extensive experimental results on facial, object, digits image and text data demonstrate that the proposed method  outperforms the state-of-the-art methods in terms of many criteria.
\end{abstract}
}
\maketitle
\IEEEdisplaynotcompsoctitleabstractindextext
\IEEEpeerreviewmaketitle

\section{Introduction}
\IEEEPARstart{B}{enefitting} from the advance of information technology, multiple views of objects can be readily acquired in many real-world scenarios, which include different kinds of features \cite{Sun2013}\cite{XuTaoXu2013}. In essence, most datasets are comprised of multiple feature sets or views. For instance, an object can be characterized by a color view and/or a shape view; an image can be depicted by different features such as color histogram and Fourier shape information, etc. These multi-view data provide more useful information, compared to single-view data, to boost clustering performance by integrating different views \cite{BickelScheffer2004,KumarRaiDaume2011}. In general, multi-view clustering \cite{BickelScheffer2004,KumarRaiDaume2011,Sun2013,XuTaoXu2013}  is superior to single-view one due to utilizing the complementary information of objects from different feature spaces.

However, a challenging problem may arise when data from different views show a large divergence, or being heterogeneous \cite{DingFu2016}.
As such, it will lead to view disagreement \cite{WangLinWuZhangZhangHuang2015} so as to fail to obtain a similarity matrix that can depict the samples within the same class. Specifically, the within-class samples across multiple views may show a lower affinity than that within the same view but from different classes \cite{DingFu2016}. In order to address this problem, a surge of methods in multi-view learning have been proposed \cite{KumarRaiDaume2011,LiuWangGaoHan2013,TzortzisLikas2012,WangNieHuang2013,WhiteYuZhangSchuurmans2012,YinWuHeWang2015}. Tzortzis \emph{et. al} \cite{TzortzisLikas2012} proposed to compute separate kernels on each view and then combined with a kernel-based method to improve clustering. To better capture the view-wise relationships among data, in work \cite{WangNieHuang2013}, a novel multi-view learning model has been presented via a joint structured sparsity-inducing norm. For exploiting the correlation consensus, a co-regularized multi-view spectral clustering \cite{WhiteYuZhangSchuurmans2012} is developed by using two co-regularization schemes. Liu \emph{et. al} \cite{LiuWangGaoHan2013} proposed a non-negative matrix factorization (NMF) based multi-view clustering algorithm via seeking for a factorization that gives compatible clustering solutions across multiple views. By taking advantage of graph Laplacian matrices \cite{YinGaoLin2016}\cite{YinGaoLinShiGuo2014} in different views, the algorithm proposed in \cite{CollinsLiuXuMukherjeeSingh2014} learns a common representation under the spectral clustering framework. Though the aforementioned methods indeed enhance the clustering
performance for multi-view data, some useful prior information within data are often ignored, such as sparsity \cite{ElhamifarVidal2013} and low-rank \cite{LiuLinYanSunMa2013}, etc. To tackle this problem, a novel pairwise sparse subspace representation model for multi-view clustering was proposed recently \cite{YinWuHeWang2015}. Ding \emph{et. al} \cite{DingFu2014} developed a robust multi-view subspace learning algorithm by seeking a common low-rank linear projection to mitigate the semantic gap among different views. Xia \emph{et. al} \cite{XiaPanDuYin2014} presented recovering a shared low-rank transition probability matrix, in favor of low-rank and sparse decomposition, and then input to the standard Markov chain method for clustering. To further mitigate the divergence between different views, Ding \emph{et. al} \cite{DingFu2016} proposed a robust multi-view subspace learning algorithm (RMSL)
through dual low-rank decompositions, which is expected to recover a low-dimensional view-invariant subspace for multi-view data. In fact, this type of subspace learning approaches aims to achieve a latent subspace shared by multiple views provided the input views are drawn from this latent subspace.

In recent years, subspace clustering has attracted considerable attentions in computer vision and machine learning communities due to its capability of clustering  data efficiently \citep{Vidal2011}. The underlying assumption is that observed data usually lie in/near some low-dimensional subspaces \cite{ParsonsHaqueLiu2004}. By constructing a pairwise similarity graph, data clustering can be readily transformed into a graph partition problem \cite{ShiMalik1997,YinGaoLinShiGuo2014,YinGaoLin2016}. The success of subspace clustering is based on a block diagonal solution that is achieved given that the objective functions satisfy some enforced block diagonal (EBD) conditions \cite{LuMinZhaoZhuHuangYan2012}. Mathematically, the objective functions are designed as a reconstruction term with different regularization, such as, either $\ell_1$-minimization (SSC) \cite{ElhamifarVidal2013}, rank minimization (LRR) \cite{LiuLinYanSunMa2013}
or $\ell_2$-regularization (LSR) \cite{LuMinZhaoZhuHuangYan2012}. Although subspace learning shows good performance in multi-view clustering, they may not fully make use of the properties of multi-view data. As discussed above, most previous methods focus on capturing only the pairwise correlations between different views, rather than the higher order correlation \cite{PengLiShao2008} underlying the multi-view data. In fact, the real world data are ubiquitously in multi-dimension, often referred to as {\it{tensors}}. Based on this observation, especially for multi-view data, omitting correlations in original spatial structure cannot result in optimal clustering performance generally. To address this issue, Zhang \emph{et. al.} \cite{ZhangFuLiuLiuCao2015} proposed a low-rank tensor constrained multi-view subspace clustering to explore the complementary information from multiple views. However, the work \cite{ZhangFuLiuLiuCao2015} cannot capture high order correlations well since it does not actually represent view data as a {\it{tensor}}.

Recently, the \textit{t-product} \cite{KilmerMehrmanKilmerMartin2011}, one type of tensor-tensor products, was introduced to provide a matrix-like multiplication for third-order tensors. The t-product shares many similar properties as the matrix product and it has become a better way of exploiting the intrinsic structure of third-order or higher order tensor \cite{KernfeldAeronKilmer2014}, against the traditional Kronecker product operator \cite{KoldaBader2009}. To perform subspace clustering on data with second-order tensor structure, i.e., images and multi-view data, conventional methods usually unfold the data or map them to vectors. Thus blind vectorizing may cause the problem of ``curse of dimensionality'' and also damage the second-order structure, like spatial information, within data. In contrast, t-product provides a novel algebraic approach for convolution operation rather than scalar multiplication \cite{KilmerMehrmanKilmerMartin2011}. Owing to this operator, a third-order tensor can be readily regarded as a ``matrix'' whose elements are n-tuples or tubes, such that the matrix data can be embedded into a vector-space-like structure \cite{KernfeldAeronKilmer2014}. To exactly recover a low-rank third-order tensor corrupted by sparse errors, most recent work \cite{LuFengChenLiuLinYan2016} studied the Tensor Robust Principal Component (TRPCA). To perform submodule clustering of multi-way data, Piao \emph{et. al} \cite{PiaoHuGaoSunLin2016} proposed a clustering method by sparse and low-rank representation using \textit{t-product}. However, this method is not developed for multi-view data, which is in favor of the linear separability assumption rather than complementary information of multi-view data. In fact, it is easier to treat multi-view data as a third-order tensor by organizing all different views of an object together, referring to Section \ref{t_MVD} for more details.

\begin{figure*}[ht]
\centering
 {\includegraphics[width= 0.8 \textwidth]{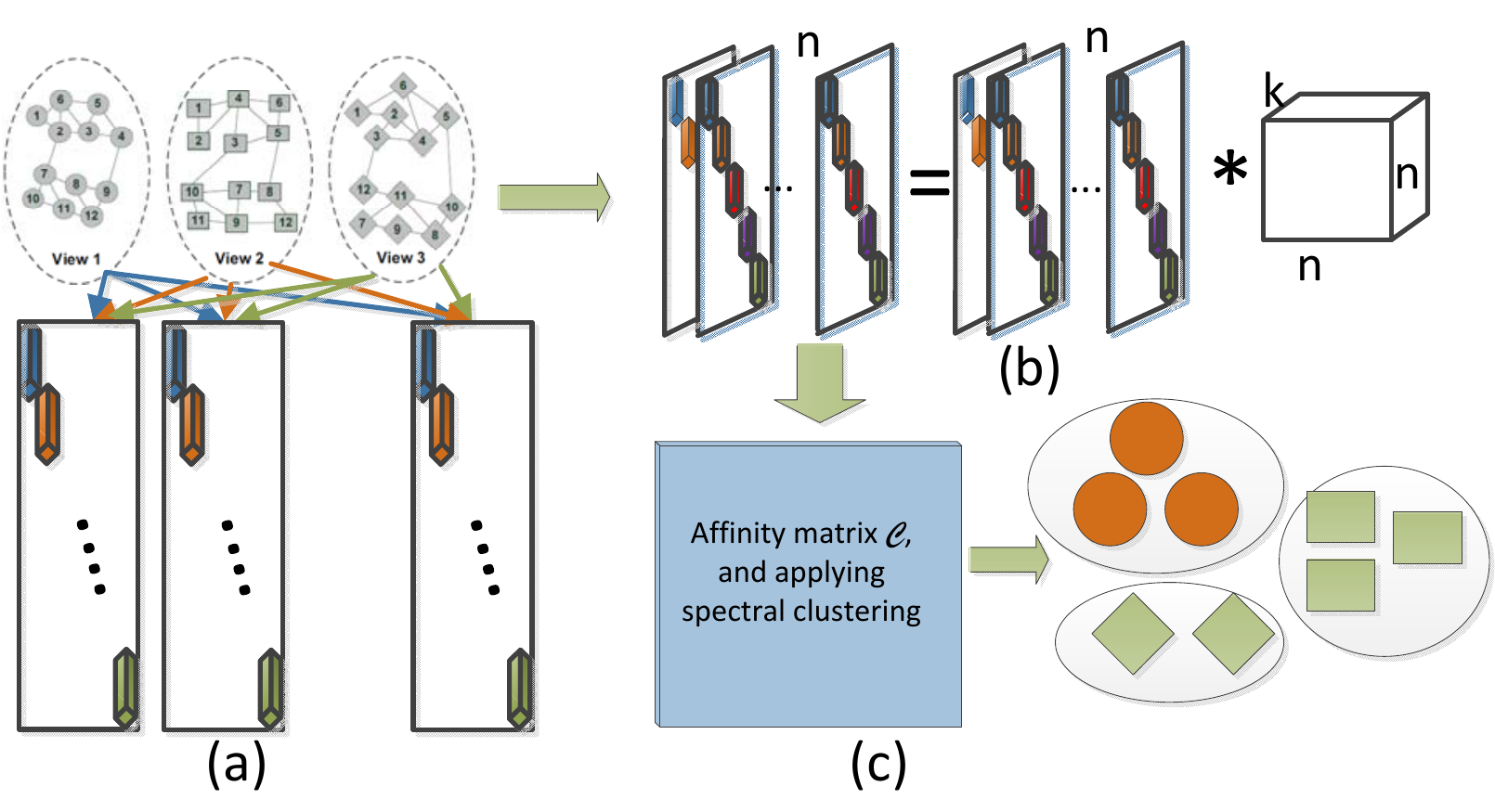}}
   \caption{Overview of the proposed framework. (a) Using third-order tensorial data to represent a multi-view dataset. Each lateral slice of tensor is formed by all views from one object where each view data is arranged at the diagonal position one-by-one. (b) By \textit{t-linear} combination, the third-order tensorial data is reconstructed by itself with sparse and low-rank penalty in \textit{self-expressive} way. (c) Based on the learned tensor coefficients, a data similarity matrix is built for multi-views, by which the spectral clustering is applied to the final separation. }\label{Overview}
\end{figure*}

Motivated by the above observations, we propose a novel low-rank multi-view clustering method  by using \textit{t-product} based-on the circular convolution in this paper. The proposed method aims to capture within-view relationships among multi-view data while respecting the feature-wise effect of each data point. By some manipulations, we can naturally transform the multiple views data of interest into a third-order tensor. In nature, the multi-view data is readily regarded as a \textit{tensor}. In what follows, we can apply the recent advance of third-order tensor algebra tools \cite{KilmerMartin2011,KilmerBramanHaoHoover2013,ZhangElyAeronHaoKilmer2013} to performing clustering or classification tasks. Specifically, each sample from different views (i.e., with $D \times k$ ) can be twisted into a third-order tensor with $D \times 1 \times k$ and all samples can be organized as a tensor with $D \times n\times k$. Then the tensorial data can be represented by the t-linear combination for data ``self-expressiveness''. The overview of our proposed method is shown in Fig. \ref{Overview}.

Our main contributions in this paper are summarized from the
following three aspects:
\begin{enumerate}
\item First, we present an innovative construction method by effectively organizing multi-view data set into third-order tensorial data. As such, multiple views can be simultaneously exploited, rather than only pairwise information.

\item More importantly, to the best of our knowledge, it is the first time to propose a low-rank multi-view clustering in third-order tensor space.
Through using \textit{t-product} based on the circular convolution operation, the multi-view data is represented by a \textit{t-linear} combination imposed by sparse and low-rank penalty using ``self-expressiveness''. Therefore, the high order structural information among all views can be efficiently explored and the underlying subspace structure within data can be also revealed.

\item We perform the proposed approach on the extensive multi-view databases, such as  facial, object, digits image and text data, to verify the effectiveness of the algorithm.
\end{enumerate}

The remainder of this paper is organized as follows. In Section
\ref{Definition}, we introduce some notations and definitions used throughout this paper. Section
\ref{Related} briefly reviews the related works. Section
\ref{Proposed} is dedicated to presenting the proposed multi-view clustering. In Section \ref{Experiment}, we present experimental results on evaluating clustering performance for several databases. Finally, Section \ref{Conclusion} concludes our paper.

\section{Notations and definitions}\label{Definition}
In this section, we would like to introduce some notations and  some relevant definitions.
Throughout this paper, we utilize calligraphy letters for tensors, e.g. $\mathcal{A} \in \mathbb{R} ^{n_1 \times n_2 \times n_3}$, bold lowercase letters for vectors, e.g. $\mathbf{a}$, uppercase for matrices, e.g. ${A}$, lowercase letters for entries, e.g. $a$,
$a_{ij}$ denotes the $(i,j)$-th entry of matrix $A$. $\| \textbf{a} \|_1 = \sum_{i} |{a}_i|$ and $\|\textbf{a} \|_2 = \sqrt{\textbf{a}^T\textbf{a}}$  are the $\ell_1$ and $\ell_2$ norms respectively, where $^T$ is the transpose operation. The matrix Frobenius norm is defined as $\left\| A \right\|_F = \sqrt{\sum_{ij }{{\left| {a_{ij} } \right|^2 }}}$. $\|A\|_*$ is the nuclear norm, defined as the sum of all singular values of $A$, which is the convex envelope of the rank operator. $\|A \|_{2,1}$ is the $\ell_{2,1}$-norm defined by $\|A \|_{2,1}= \sum_j{\sqrt{\sum_i a_{ij}^2}}$.

We also use Matlab notation to denote the elements in tensors. Specifically, $\mathcal{A}(:,:,i)$, $\mathcal{A}(:,i,:)$ and $\mathcal{A}(i,:,:)$ are represented by the $i$-th frontal, lateral and horizontal slice, respectively. $\mathcal{A}(:,i,j)$, $\mathcal{A}(i,:,j)$ and $\mathcal{A}(i,j,:)$ denote the mode-1, mode-2 and mode-3 fiber, respectively. We denote $\hat{\mathcal{A}}$ the Discrete Fourier Transform (DFT) along mode-3 for a third-order tensor $\mathcal{A}$, i.e., $\hat{\mathcal{A}} = \text{fft}(\mathcal{A},[\;],3)$.  Similarly, $\mathcal{A}$ can be computed by $\hat{\mathcal{A}}$ via ifft$(\hat{\mathcal{A}},[\;],3)$, i.e., using inverse fast Fourier transform (FFT). ${A}^{(i)}$ and $\hat{{A}}^{(i)}$ denote the $i$-th frontal slice of $\mathcal{A}$ and $\hat{\mathcal{A}}$, respectively.
We give the following definitions, similar to those in \cite{KernfeldAeronKilmer2014}.
\begin{defn}[block diagonal operation (bdiag)\cite{KilmerBramanHaoHoover2013}] For $\mathcal{A}$, its  block diagonal matrix is formed by its frontal slice with each block on diagonal.
\begin{equation}
\begin{aligned}
&\text{bdiag}(\mathcal{A})=\left[\begin{array}{cccc}
    {A}^{(1)} &&&  \\
    & {A}^{(2)} &&  \\
    && \ddots &  \\
    &&& {A}^{(n_3)} \\
  \end{array}\right].
\end{aligned}
\label{eqt_bdiag}
\end{equation}
\end{defn}

\begin{defn}[block circulant operation (bcirc)] For $\mathcal{A}$, its block diagonal matrix is defined as following.
\begin{equation}
\begin{aligned}
\text{bcirc}(\mathcal{A})=\left[\begin{array}{cccc}
    {A}^{(1)} & {A}^{(n_3)}     & \cdots & {A}^{(2)} \\
    {A}^{(2)} & {A}^{(1)}       & \cdots & {A}^{(3)} \\
    \vdots            & \vdots                  & \ddots & \vdots            \\
    {A}^{(n_3)} & {A}^{(n_3-1)} & \cdots & {A}^{(1)} \\
  \end{array}\right].
\end{aligned}
\label{eqt_bcirc}
\end{equation}
\end{defn}

\begin{defn}[unfold and fold operation] Unfold and fold operations are defined as following.
\begin{equation}
\begin{aligned}
\text{unfold}(\mathcal{A})=\left[\begin{array}{c}
    {A}^{(1)}   \\
    {A}^{(2)}   \\
    \vdots              \\
    {A}^{(n_3)} \\
  \end{array}\right],
\quad \text{fold}(\text{unfold}(\mathcal{A}))=\mathcal{A}.
\end{aligned}
\label{eqt_unfold_fold}
\end{equation}
\end{defn}

\begin{defn}[t-product]
Let $\mathcal{A}\in\mathbb{R}^{n_1\times n_2\times n_3}$ and $\mathcal{B}\in\mathbb{R}^{n_2\times n_4\times n_3}$, then the \textit{t-product} of $\mathcal{A}$ and $\mathcal{B}$ is defined by $\mathcal{C}\in\mathbb{R}^{n_1\times n_4\times n_3}$ as follows:
\begin{equation}
\mathcal{C}=\mathcal{A}*\mathcal{B}=\rm{fold}(\rm{bcirc}(\mathcal{A})\rm{unfold}(\mathcal{B})).
\label{eqt_tproduct}
\end{equation}
In fact, the t-product $*$ is also called  the the circular convolution operation \cite{KilmerMehrmanKilmerMartin2011}.
\end{defn}

Note that a third-order tensor $\mathcal{A}$ can be seen as an $n_1 \times n_2$ matrix with each entry as a tube lying in the mode-3. Then, the t-product operation, analogous to matrix-matrix product, is a useful generalization of matrix multiplication for tensors \cite{KilmerBramanHaoHoover2013}, except that the circular convolution replaces the product operation between the elements. Note that the t-product reduces to the standard matrix-matrix product in the case of $n_3 = 1$. Moreover, due to its superiority in generalization of matrix multiplication, the t-product has been exploited in third or higher order tensors analysis \cite{KilmerMartin2011,KilmerBramanHaoHoover2013,ZhangElyAeronHaoKilmer2013}. Based on this observation, we can efficiently exploit the linear algebra for tensors with t-product operation.

\begin{defn}[Tensor multi-rank] The multi-rank of $\mathcal{A} \in \mathbb{R}^{n_1 \times n_2 \times n_3}$ is a vector $p \in \mathbb{R}^{n_3}$
with the $i$-th element equal to the rank of the $i$-th frontal slice of $\hat{\mathcal{A}}$.
\end{defn}

\begin{defn}[Tensor nuclear norm] The tensor nuclear norm (TNN), denoted by $\|\mathcal{A} \|_{TNN}$, is  defined as the sum of the singular values of all the frontal slices of $\hat{\mathcal{A}}$, and it is the tightest convex relaxation to $\ell_1$ norm of the tensor multi-rank. That is, $\|\mathcal{A} \|_{TNN} = \sum_{i}\|\mathcal{A}(:,:,i)\|_*$.
\end{defn}

\begin{defn}[F1 norm] The F1 norm of a tensor $\mathcal{A}$ is defined by $\|\mathcal{A}\|_{F1}=\sum_{i,j}\|\mathcal{A}(i,j,:)\|_F$.
\end{defn}

\begin{defn}[FF1 norm] The FF1 norm of a tensor $\mathcal{A}$ is defined by $\|\mathcal{A}\|_{FF1}=\sum_{i}\|\mathcal{A}(i,:,:)\|_F$.
\end{defn}

\begin{defn}[Frobenius norm] The Frobenius norm of a tensor $\mathcal{A}$ is defined by $\|\mathcal{A}\|_F= \sqrt{\sum_{i,j,k}\mathcal{A}(i,j,k)^2}$.
\end{defn}

\section{Related work}\label{Related}
Before presenting our proposed method, we briefly review the background of our proposed method, which includes multi-view clustering, low-rank clustering and \textit{t-linear} combination.
\subsection{Sparse and Low-Rank Subspace Clustering}
Sparse and low rank information of the latent group structure have been exploited for subspace clustering successfully in recent years \cite{LiuLinYanSunMa2013,VidalFavaro2014,YinGaoGuo2015,YinGaoLin2016,YinGaoLinShiGuo2014}.
The underlying assumption is that data are drawn from a mixture of several low-dimensional subspaces approximately. Given a set of data points, each of them in a union of subspaces can be represented as a linear combination of points belonging to the same subspace via {\it{self-expressive}}. Specifically, for data $X = [\mathbf{x}_1, \mathbf{x}_2,..., \mathbf{x}_n], \mathbf x_i \in \mathbb{R}^d$ sampled from a union of multiple subspaces $\bigcup^M_{m=1}\mathcal{S}_m$, where $\mathcal{S}_1$, $\mathcal{S}_2$, ..., $\mathcal{S}_M$ are low-dimensional subspaces. The sparse and low-rank subspace clustering \cite{ZhuangGaoLinMaZhangYu2012} focuses on solving the following optimization problem,
\begin{equation}
\begin{aligned}
\mathop{\min}\limits_{C,E} ~ &\|C \|_* +  \lambda \left\| C \right\|_1+ \beta \|E \|_{2,1},  \\&\textrm{ s.t. } ~X = XC +E, C \geq 0.
\label{LRSC}
\end{aligned}
\end{equation}
where $C$ is the representation matrix and $E$ is the representation error. The $\ell_{2,1}$ is used in \eqref{LRSC} to cope with the gross loss across different data cases.   $\lambda$ and $\beta$ are the penalty parameter balancing the low-rank constraint, the sparsity term and the gross error term, respectively. In this model, both sparsity and lowest rank criteria, as well as a non-negative constraint, are all imposed. By imposing low rankness criterion, the global structure of data X is better captured, while the sparsity criterion can further encourage the local structure of each data vector \cite{ZhuangGaoLinMaZhangYu2012}.

In general, there are two explanations for $C$ based on this model. Firstly, the $(ij)$-th element of $C$, i.e. $c_{ij}$, reflects the ''similarity" between the pair $\mathbf x_i$ and $\mathbf x_j$. Hence $C$ is sometimes called affinity matrix; Secondly,  the $i$-th column  of $C$, i.e. $\mathbf c_i$, as a ``better'' representation of  $\mathbf x_i$  such that the desired pattern, say subspace structure, is more prominent.

\subsection{Multi-view Clustering}
To sufficiently exploit the complementary information of objects among multiple views, a surge of approaches have been proposed recently. In general, the existing methods for multi-view clustering can be roughly grouped into three categories. The first class aims at seeking some shared representation via incorporating the information of different views. That is, it maximizes the mutual agreement on two distinct views of the data\cite{BickelScheffer2004,KumarDaume2011,YuKrishnapuramRosalesRao2011}. For example, Kumar \emph{et. al.} \cite{KumarDaume2011} first proposed the co-training spectral clustering algorithm for multi-view data. Under the assumption that view data are generated by a mixture model, Bickel \emph{et. al.} \cite{BickelScheffer2004} applied expectation-maximization (EM) in each view and then clustered the data into subsets with high probability. The second one is called ensemble clustering, or late fusion \cite{TzortzisLikas2012}.

The core idea behind the aforementioned methods is to utilize kernels that naturally correspond to each single view and integrate kernels either linearly or non-linearly to get a final grouping output \cite{GreeneCunningham2009,TzortzisLikas2012}. Tzortzis \emph{et. al} \cite{TzortzisLikas2012} proposed  computing separate kernels on each view and then combined with a kernel-based method to improve clustering. A matrix factorization based method is presented to group the clusters obtained from each view \cite{GreeneCunningham2009}, which is termed as subspace learning based methods \cite{DingFu2014,LiuWangGaoHan2013,XiaPanDuYin2014,YinWuHeWang2015,ZhangFuLiuLiuCao2015}. Based on the assumption that each input view is generated from a latent subspace, it focuses on achieving this latent subspace shared by multiple views. Recent works \cite{ElhamifarVidal2013,LiuLinYanSunMa2013,LuMinZhaoZhuHuangYan2012} show that some useful prior knowledge, such as sparse or low-rank information, can help capture the latent group structure to improve clustering performance.

Motivated by this observation, in this paper, we aim to take advantage of the higher order correlation underlying the multi-view data in a third-order tensor space.

\subsection{\textit{t-linear} Combination}
To better capture the higher order correlation among data, especially for original spatial structure, it is desirable that third-order tensors can be operated like matrices  using linear algebra tools. Although many tensor decompositions \cite{KoldaBader2009}, such as CANDECOMP/PARAFAC (CP), Tucker and Higher-Order SVD \cite{LathauwerMoorVandewalle2000}, facilitate the
linear algebra tools to multi-linear context, this extension cannot be understood well for third-order tensors. To address this problem, Kilmer \emph{et. al.} \cite{KilmerBramanHaoHoover2013} recently presented t-product to define a matrix-like multiplication for third-order tensors. Given a matrix with size of $m \times n$, one can \textit{twist} it into a ``page'' and then form an $m \times 1 \times n$ third-order tensor (``oriented matrices'')%,illustrated in Fig. \ref{t_twist}
. 
Note that an $m \times 1 \times n$ third-order tensor is really a tensor rather than a matrix. In fact, the tensor with size of  $m \times 1 \times n$ can be regarded as a vector of length $m$, where each element is an $1 \times 1 \times n$ tube fiber (called tube fiber as usual in tensor literature). Benefit from t-product \cite{KilmerBramanHaoHoover2013}, one can multiply two tube fibers, and then we can present ``linear'' combinations of oriented matrices \cite{KernfeldAeronKilmer2014}. That is, the operation is defined by {\it{t-linear}} combination, where the coefficients are tube fibers, and not scalars. %The illustration is given in Fig. \ref{t_linear} where $*$ denotes the circular convolution. 
Under this definition, a tensor $\mathcal{A}$ with $m \times 1 \times n$ is represented as a combinations of a tensor $\mathcal{X}$(size of $m \times m \times n$) with $\mathcal{B}$ (size of $m \times 1 \times n$). For more details of t-product, please refer to \cite{KilmerBramanHaoHoover2013}.

%\begin{figure}[ht]
%\centering
% {\includegraphics[width=0.3 \textwidth]{ttwist.pdf}}
%   \caption{Operation between matrix and third-order tensor. A $m \times n$ matrix(left) can be transformed into a tensor(right) with $m \times 1 \times n$ by twist operation, and vice versa, by squeeze operation. }\label{t_twist}
%\end{figure}
%\begin{figure}[ht]
%\centering
% {\includegraphics[width=0.4 \textwidth]{ttlinear.pdf}}
%   \caption{An illustration of a third-order tensor $\mathcal{A}$ is linearly combined by the coefficients $\mathcal{B}$ under a basis $\mathcal{X}$. }\label{t_linear}
%\end{figure}

\section{Proposed method}\label{Proposed}
To efficiently incorporate the clustering results from different views, we first organize each data point by a third-order tensor with all views information, in Section \ref{t_MVD}. As a result, one can maximize the agreement on multiple distinct view while recognizing the complementary information contained in each view. Then, in Section \ref{OurMethod}, we propose a sparse and low-rank clustering method for multi-view data in third-order tensor space, followed by an optimization via the alternating direction method of multipliers (ADMM) in Section \ref{Optimizing}. Subspace clustering for multi-view data is performed through spectral clustering in Section \ref{SC}. Finally, convergence and computational complexity analysis of the proposed algorithm are discussed in Sections \ref{CompConv}.

\subsection{Multi-view data represented by third-order tensor}\label{t_MVD}
Given a multi-view data set $X_v \in \mathbb{R}^{d_v \times n}$, which includes the features of the $v$-th view ($v = 1, 2, ..., k$, totally $k$ views). To integrate all views for the $i$-th object ($i = 1, 2, ..., n$), we build a matrix $\mathbf{M}_i \in \mathbb{R}^{D \times k}$, $D= \sum_v d_v$, where its diagonal position are composed of each view data.
%The building process is illustrated in Fig. \ref{tconstr}. 
That is, the $j$-th column of $\mathbf{M}_i$ consists of the $j$-th view data. By this organizing, the set of $\{\mathbf{M}_i\}_{i=1}^n$ is able to convey the complementary information across multiple views without enforcing clustering agreement among distinct views. Furthermore, this leads to an union of different views whilst respecting each individual view data. Through using \textit{twist} manipulation, the multi-view data for the $i$-th object is easily transformed into a third-order tensor space, i.e., $\mathcal{M}_i \in \mathbb{R}^{D \times 1 \times  k}$. Collecting all $\{\mathcal{M}_i\}$ along the second mode, we can obtain a tensor $\mathcal{X} \in \mathbb{R}^{D \times n \times  k}$. As a consequence, the proposed clustering method can be effectively applied to this third-order tensor such that the high order correlation can be exploited by using all views simultaneously.

\subsection{Sparse and low-rank clustering in third-order tensor space}\label{OurMethod}
Given multi-view data $\mathcal{X} \in \mathbb{R}^{D \times n \times k}$, it is crucial to find a method to effectively represent the data, in {\it{self-expressive}} way, for clustering task. In literature, a lot of work have been presented for \textit{matrix-data} in order to discover the pairwise correlations between different views \cite{DingFu2016,KumarRaiDaume2011,TzortzisLikas2012,WangLinWuZhangZhangHuang2015,WhiteYuZhangSchuurmans2012,YinWuHeWang2015}.  To generalize the clustering methods for the matrix case to the one for third or higher order tensorial cases, Kernfeld \emph{et. al.} \cite{KernfeldAeronKilmer2014} recently proposed a sparse submodule clustering method(termed as SSmC), which can be formulated as follows.
\begin{equation}
\begin{aligned}
\min_{\mathcal{C}}\quad&\|\mathcal{C}\|_{F1}+ \lambda_1\| \mathcal{C}\|_{FF1}+ \frac{\lambda_2}{2} \|\mathcal{X- X*C}\|_F^2, \\~\text{s.t.}\quad&\mathcal{C}(i,i,v)=0, i=1,2,...,n, ~v= 1,2,...,k .
\end{aligned}
\end{equation}
where $\mathcal{C}$ is the representation tensor, $\lambda_1$ and $\lambda_2$ are the balance parameters.

However, this model cannot be applicable to the multi-view data clustering directly, due to the consensus principle in multi-view data \cite{XuTaoXu2013}. In addition, the success of low-rank regularizer has been widely witnessed in many work \cite{DingFu2014,YinGaoGuo2015,YinGaoLin2016,YinGaoLinShiGuo2014,ZhangFuLiuLiuCao2015}. Thus, in this section, we propose to seek a most sparsity and lowest-rank representation of multi-view data by exploiting the {\it{self-expressive}} property%, illustrated by Fig. \ref{model}
. Mathematically, it can be formulated as follows,
\begin{equation}
\begin{aligned}
\min_{\mathcal{C}}\quad& \alpha\|\mathcal{C}\|_{F1}+ \lambda \|\mathcal{C}\|_{TNN}+ \frac1{2} \|\mathcal{X- X*C}\|_F^2 \\ \quad& + \frac{\beta}{2}\sum\limits_{1\leq i,j\leq k, i\neq j}\|\mathcal{C}(:,:,i)- \mathcal{C}(:,:,j)\|_F^2 .\label{proposed}
\end{aligned}
\end{equation}
where $\mathcal{C} \in  \mathbb{R}^{n \times n \times k}$ denotes the representation tensor utilized to induce the following ``affinity'' matrix. $\|\cdot\|_{F1}$ and $\|\cdot\|_{TNN}$ are the tensor sparse and nuclear norm, respectively, as defined in Section \ref{Definition}. Based on these two norms, the first and second terms of the objective function aim to induce sparse and lowest-rank coefficients. The third term fits the representation errors in third-order tensor space by using \textit{t-product}. Finally, the last term is imposed for multi-view data in particular, which encourages the consensus clustering via forcing all the lowest-rank coefficients close in all the views.  For ease of numeric implementation, we here employ the Frobenius norm rather than $\ell_1$ norm.

\subsection{Optimization via ADMM}\label{Optimizing}
Variable $\mathcal{C}$ appears in three terms in the objective function \eqref{proposed}. To decouple them, we introduce two variables $\mathcal{Y} = \mathcal{C}$ and $\mathcal{Z} = \mathcal{C}$. Then, we have the following problem such that the standard ADMM\cite{WenGoldfarbYin2010} can be efficiently applied to.
\begin{equation}
\begin{aligned}
\min_{\mathcal{C,Y, Z}}\quad& \alpha\|\mathcal{Y}\|_{F1}+ \lambda \|\mathcal{Z}\|_{TNN}+ \frac1{2} \|\mathcal{X- X*C}\|_F^2 \\ \quad& + \frac{\beta}{2}\sum\limits_{1\leq i,j\leq k, i\neq j}\|\mathcal{C}(:,:,i)- \mathcal{C}(:,:,j)\|_F^2 , \\
&\text{s.t.}, ~ \mathcal{Y= C, Z = C}.
\end{aligned}
\end{equation}
Its augmented Lagrangian formulation is formulated as follows,
\begin{equation}
\begin{aligned}
\argmin_{\mathcal{C,Y,Z}}\quad& \alpha\|\mathcal{Y}\|_{F1}+ \lambda \|\mathcal{Z}\|_{TNN}+ \frac1{2} \|\mathcal{X- X*C}\|_F^2 \\
\quad& + \frac{\beta}{2}\sum\limits_{1\leq i,j\leq k, i\neq j}\|\mathcal{C}(:,:,i)- \mathcal{C}(:,:,j)\|_F^2 \\
\quad&+ \langle \mathcal{G}_1, \mathcal{Y- C}\rangle + \langle \mathcal{G}_2,\mathcal{Z -C} \rangle\\
\quad& + \frac{\rho}{2}(\| \mathcal{Y- C}\|_F^2+ \| \mathcal{Z- C}\|_F^2).\label{AugProb}
\end{aligned}
\end{equation}
where $\mathcal{G}_1$ and $\mathcal{G}_2$ are Lagrange multipliers, and $\rho > 0$ is a
penalty parameter. As convolution-multiplication properties, this problem can be computed efficiently in the Fourier domain. Then, the procedure of solving \eqref{AugProb} with ADMM is defined as follows,
\begin{enumerate}
\item Updating $\mathcal{Z}$ by
\begin{equation}
\begin{aligned}
\argmin_{\mathcal{Z}}\quad&  \lambda \|\mathcal{Z}\|_{TNN} + \langle \mathcal{G}_2,\mathcal{Z -C} \rangle+ \frac{\rho}{2}\| \mathcal{Z- C}\|_F^2 \\
\quad& = \lambda \|\mathcal{Z}\|_{TNN} +  \frac{\rho}{2}\| \mathcal{Z- C} + \frac1{\rho}\mathcal{G}_2\|_F^2.
\end{aligned}
\end{equation}
From the frontal side, e.g., ${C}_i= \mathcal{C}(:,:,i)$, $\mathcal{Z}$ can be optimized slice-by-slice. That is, the sub-problem is equivalent to solving
\begin{equation}
\begin{aligned}
{Z}_i^{t+1} = \argmin_{{Z}_i}\quad&  \lambda \|{Z}_i\|_* +  \frac{\rho}{2}\| {Z_i- C_i} + \frac1{\rho}{G}_2^i\|_F^2. \label{SolveZ}
\end{aligned}
\end{equation}
which has a closed-form solution by using   the Singular Value Thresholding (SVT) operator \citep{CaiCandesShen2008}.

\item Updating $\mathcal{Y}$ by
\begin{equation}
\begin{aligned}
\argmin_{\mathcal{Y}}\quad& \alpha\|\mathcal{Y}\|_{F1}+  \langle \mathcal{G}_1, \mathcal{Y- C}\rangle +  \frac{\rho}{2}\| \mathcal{Y- C}\|_F^2.
\end{aligned}
\end{equation}
Similarly, $\mathcal{Y}$ can be efficiently solved from the third mode fiber-by-fiber. That is,
\begin{equation}
\mathcal{Y}(i,j,:)^{t+1} = \frac{\|\mathcal{A}(i,j,:)\|_F-\frac{\alpha}{\rho}}{\|\mathcal{A}(i,j,:)\|_F} \mathcal{A}(i,j,:).
\label{SolveY}
\end{equation}
where $\mathcal{A}=\mathcal{C}-\frac1{\rho}{\mathcal{G}_1}$.

\item Updating $\mathcal{C}$ by
\begin{equation}
\begin{aligned}
\argmin_{\mathcal{C}}
\quad& \frac1{2} \|\mathcal{X- X*C}\|_F^2+ \langle \mathcal{G}_1, \mathcal{Y- C}\rangle +\langle \mathcal{G}_2,\mathcal{Z -C} \rangle\\
\quad& + \frac{\beta}{2}\sum\limits_{1\leq i,j\leq k, i\neq j}\|\mathcal{C}(:,:,i)- \mathcal{C}(:,:,j)\|_F^2 \\
\quad& + \frac{\rho}{2}(\| \mathcal{Y- C}\|_F^2+ \| \mathcal{Z- C}\|_F^2). \label{SubPC}
\end{aligned}
\end{equation}

By letting $\mathcal{P}_1= \mathcal{Z}+ \frac1{\rho}\mathcal{G }_1$,  $\mathcal{P}_2= \mathcal{Y}+ \frac1{\rho}\mathcal{G }_2$ and applying FFT, we have the following equivalent problem\footnote{Note that, roughly, the sum of the square of a function is equal to the sum of the square of its transform, according to Parseval's theorem\cite{Arfken1985}. },
\begin{equation}
\begin{aligned}
\argmin_{\hat{\mathcal{C}}, {Q}_i}
\quad& \frac1{2} \| \hat{\mathcal{X}}- \hat{\mathcal{X}} \odot \hat{\mathcal{C}}\|_F^2+ \frac{\beta}{2}\sum\limits_{i,j}\|\hat{\mathcal{C}}(:,:,i)-  \hat{\mathcal{C}}(:,:,j)\|_F^2\\
\quad& +\frac{\rho}{2}(\| \hat{\mathcal{C}}- \hat{\mathcal{P}}_1\|_F^2+ \| \hat{\mathcal{C}}- \hat{\mathcal{P}}_2\|_F^2).
\end{aligned}
\end{equation}
where $\odot$ denotes the point-wise multiplication. That is, $\hat{\mathcal{X}} \odot \hat{\mathcal{C}}$ is an array resulting from point-wise multiplication. Then, we can optimize the problem slice-by-slice from the frontal side, i.e.,
\begin{equation}
\begin{aligned}
\argmin_{\hat{{C}}^{(i)}}
\quad& \frac1{2} \| \hat{{X}}^{(i)}- \hat{{X}}^{(i)} \hat{{C}}^{(i)}\|_F^2+ \frac{\beta}{2}\sum\limits_{i,j,i\neq j}\|\hat{{C}}^{(i)}- \hat{{C}}^{(j)}\|_F^2\\
\quad& +\frac{\rho}{2}(\| \hat{{C}}^{(i)}- \hat{{P}}_1^{(i)}\|_F^2+ \| \hat{{C}}^{(i)}- \hat{{P}}_2^{(i)}\|_F^2).\label{SubPC1}
\end{aligned}
\end{equation}

The sub-problem \eqref{SubPC1} is non-separable w.r.t. $\hat{{C}}^{(i)}$, however, thus it has to be reformulated as an equivalent problem with separable objective. Therefore, an auxiliary variable, named ${Q}_i$, is introduced. Then,
\begin{equation}
\begin{aligned}
\argmin_{\hat{{C}}^{(i)}, {Q}_i}
\quad& \frac1{2} \| \hat{{X}}^{(i)}- \hat{{X}}^{(i)} \hat{{C}}^{(i)}\|_F^2+ \frac{\beta}{2}\sum\limits_{i,j,i\neq j}\|{Q}_i- \hat{{C}}^{(j)}\|_F^2\\
\quad& +\frac{\rho}{2}(\| \hat{{C}}^{(i)}- \hat{{P}}_1^{(i)}\|_F^2+ \| \hat{{C}}^{(i)}- \hat{{P}}_2^{(i)}\|_F^2),\\ \quad& \text{s.t.}, ~{Q}_i= \hat{{C}}^{(i)}.
\end{aligned}
\end{equation}

Next, the details for alternatively updating these two blocks are given.
\begin{itemize}
\item Update $\hat{{C}}^{(i)}(i= 1,2, ..., k)$. Each $\hat{{C}}^{(i)}$ can be updated independently by,
\begin{equation}
\begin{aligned}
\argmin_{\hat{{C}}^{(i)}}
\quad& \| \hat{{X}}^{(i)}- \hat{{X}}^{(i)} \hat{{C}}^{(i)}\|_F^2 + \beta \sum\limits_{i,j}\|{Q}_j- \hat{{C}}^{(i)}\|_F^2\\
\quad& +{\rho}(\| \hat{{C}}^{(i)} - \hat{{P}}_1^{(i)}\|_F^2+ \| \hat{{C}}^{(i)}- \hat{{P}}_2^{(i)}\|_F^2)\\
\quad&  +\frac{\tau}{2} \|{Q}_j- \hat{{C}}^{(i)}+ \frac1{\tau} W_i \|_F^2 .
\end{aligned}
\end{equation}
Equivalently,
\begin{equation}
\begin{aligned}
\argmin_{\hat{{C}}^{(i)}}
{\rho}(\| \hat{{C}}^{(i)}- \hat{{P}}_1^{(i)}\|_F^2+ \| \hat{{C}}^{(i)}- \hat{{P}}_2^{(i)}\|_F^2)\\
 + \frac1{2}\beta (k-1) \|\hat{{C}}^{(i)} -\frac1{k-1} \sum\limits_{j, j\neq i}{Q}_j \|_F^2 \\+ \| \hat{{X}}_i- \hat{{X}}_i \hat{{C}}^{(i)}\|_F^2 +\frac{\tau}{2} \|{Q}_i- \hat{{C}}^{(i)}+ \frac1{\tau} W_i \|_F^2 .
\end{aligned}
\end{equation}
Taking derivation w.r.t. $\hat{{C}}^{(i)}$ and letting it be zeros, we have,
\begin{equation}
\begin{aligned}
&((\beta (k-1)+ \tau+ 2\rho)\mathbf{I}+ (\hat{{X}}^{(i)})^T\hat{{X}}^{(i)}  )(\hat{{C}}^{(i)})^* = \\
&\beta \sum\limits_{j, j\neq i}{Q}_j+ \tau(Q_i+ \frac1{\tau} W_i) +(\hat{{X}}^{(i)})^T\hat{{X}}^{(i)} \\
&+ \rho(\hat{{P}}_1^{(i)} +\hat{{P}}_2^{(i)}) .
\end{aligned}
\end{equation}
where $\mathbf{I}$ is an identity matrix.

\item Update $Q_i(i= 1,2, ..., k)$
\begin{equation}
\begin{aligned}
\argmin_{Q_i}
\frac{\beta}{2} \sum\limits_{i,j\neq i}\|{Q}_i- \hat{C}^{(j)}\|_F^2 \\
+\frac{\tau}{2} \|{Q}_i- \hat{C}^{(i)}+ \frac1{\tau} W_i \|_F^2 .
\end{aligned}
\end{equation}

Similarly, taking derivation w.r.t. $Q_i$ and letting it be zeros, we have,
\begin{equation}
\begin{aligned}
(\beta (k-1) + \tau) Q_i^* = \beta \sum\limits_{j, j\neq i}\hat{C}^{(j)} +\tau\hat{C}^{(i)} - W_i. \label{SolveQ}
\end{aligned}
\end{equation}

\item Update $W_i$
\begin{equation}
\begin{aligned}
W_i = W_i + \tau(Q_i^* - (\hat{{C}}^{(i)})^* ).
\end{aligned}
\end{equation}
\end{itemize}

\item Updating $\mathcal{G}_1$ and $\mathcal{G}_2$
\begin{equation}
\begin{aligned}
 \mathcal{G}_1=  \mathcal{G}_1+ \mu (\mathcal{Y -C}),\\
 \mathcal{G}_2=  \mathcal{G}_2+ \mu (\mathcal{Z -C}),\\
 \rho = \min \left(  \rho_{\textrm{max}} , \mu \rho \right). \label{UpdateGmu}
\end{aligned}
\end{equation}
\end{enumerate}

The whole procedure of ADMM for solving \eqref{AugProb} is summarized in Algorithm \ref{Alg1}. The stopping criterion is given by the following condition in the algorithm.
\begin{equation}
\begin{aligned}
\rm{max}\left\{
\begin{array}{c}
\frac1{\|\mathcal{X}\|_F}\|\mathcal{Z}^{t+1}-\mathcal{C}^{t+1}\|_F, \frac1{\|\mathcal{X}\|_F}\|\mathcal{Y}^{t+1}-\mathcal{C}^{t+1}\|_F,\\
\frac1{\|\mathcal{Z}^t\|_F}\|\mathcal{Z}^{t+1}-\mathcal{Z}^{t}\|_F, \frac1{\|\mathcal{Y}^t\|_F}\|\mathcal{Y}^{t+1}-\mathcal{Y}^{t}\|_F,\\
\frac1{\|\mathcal{C}^t\|_F}\|\mathcal{C}^{t+1}-\mathcal{C}^{t}\|_F
  \end{array}
\right\}\leq\varepsilon.
\end{aligned}
\label{ConvCondition}
\end{equation}

\begin{algorithm}
\caption{Solving problem\eqref{AugProb} via ADMM.}
\SetKwData{Index}{Index}
\KwIn{$\mathcal{X}$ , $\lambda$, $\alpha$ and $\beta$.}
\textbf{Initialization:} $\mathcal{C}^0= \mathcal{Y}^0= \mathcal{Z}^0= 0$, $\mathcal{G}_1^0=\mathcal{G}_2^0= 0$, $\rho =\tau = 0.01$, $\mu= 1.9$.
\BlankLine
\textbf{While} not converged $\left({\mathit{t}=0,1,...} \right)$ \textbf{do}
\begin{enumerate}
\item  Update $\mathcal{Z}^{t+1}$ according to \eqref{SolveZ};
\item  Update $\mathcal{Y}^{t+1}$ according to \eqref{SolveY};
\item  Update $\mathcal{C}^{t+1}$ according to \eqref{SubPC};
%\item  Update $\mathcal{Q}^{t+1}$ according to \eqref{SolveQ};
\item  Update $\mathcal{G}_1$ , $\mathcal{G}_2$ and  $\rho$ using \eqref{UpdateGmu};
\item   Check convergence:  If the condition defined by \eqref{ConvCondition} is satisfied, then break.
\end{enumerate}
\textbf{End while}\\
\KwOut{$\mathcal{C}^*$}\label{Alg1}
\end{algorithm}

\subsection{Subspace Clustering for Multi-View Data}\label{SC}
As discussed earlier, in fact, $\mathcal{C}$ can be regarded as a new representation learned from multi-view data. After solving problem \eqref{AugProb}, the next step is to segment $\mathcal{C}$ to find the final subspace clusters. For $\mathcal{C}$, it contains $k$ affinity matrices corresponding to each view, from the frontal side. However, how to effectively combine these information is not a trivial issue. Considering the superiority of the work \cite{XiaPanDuYin2014}, here we adopt the transition probability matrix to achieve the final cluster result similarly. Specifically, we first recover the latent transition probability matrix, utilizing $\mathcal{C}$ from all views, by a decomposition method. Then the latent transition matrix will be used as input to the standard Markov chain method to separate the data into clusters \cite{XiaPanDuYin2014}. For computational complexity, we are in favor of $\ell_{2,1}$ norm rather than nuclear norm on optimizing the transition matrix. We call this algorithm Subspace Clustering for Multi-View data in third-order Tensor space (SCMV-3DT for short) and it is outlined in Algorithm \ref{Alg2}.
\begin{algorithm}
\caption{Subspace Clustering for Multi-View Data in Third-order Tensor Space. }
\SetKwData{Index}{Index}
\KwIn{ $\mathcal{X}$ ,  $\lambda$, $\alpha$ and $\beta$.}
\BlankLine
\textbf{Steps:}
\begin{enumerate}
\item Solve \eqref{AugProb} by ADMM explained in Section \ref{Optimizing},
and obtain the optimal solution $ \mathcal{C}^* $.
\item Similar to work\cite{XiaPanDuYin2014}, compute the latent transition probability matrix by $ \mathcal{C}^* $, and input to the standard Markov chain method to separate the data.
\end{enumerate}
\KwOut{the clustering solution {$\mathcal W$}.}
\label{Alg2}
\end{algorithm}

\subsection{Convergence and Complexity Analysis} \label{CompConv}
As problem \eqref{proposed} is convex, the algorithm via ADMM is guaranteed to converge at the rate of $\mathcal{O}(1/t_1)$\cite{WenGoldfarbYin2010}, where $t_1$ is the number of iterations.

The proposed algorithm consists of three steps involving in iteratively updating $\mathcal{Z}$, $\mathcal{Y}$ and $\mathcal{C}$, until the convergence condition is met. The time complexity for each update is listed in Table \ref{CompList}. From the table, we can see how our algorithm is related to the size of multi-view data.

\begin{table*}
\caption{Time complexity analysis of the proposed algorithm, where $t_2$ is the iteration times for solving sub-problem\eqref{SubPC} and $r$ is the lowest rank for $\mathcal{Z}$ that can be obtained by our algorithm.}\label{CompList}
\begin{center}
\begin{tabular}{|l|c|c|c|c|}
\hline
\textit{Algorithm} & Update $\mathcal{Z}$ & Update $\mathcal{Y}$ & Update $\mathcal{C}$ & total time complexity \\
\hline
SCMV-3DT &$\mathcal{O}(krn^2)$ & $\mathcal{O}(kn^2)$  & $\mathcal{O}(\text{max}(n^2D,n^3)k)$ & $\mathcal{O}(t_1(krn^2+t_2\text{max}(n^2D,n^3)k))$ \\
\hline
\end{tabular}
\end{center}
\end{table*}

\section{Experimental Results}\label{Experiment}
In order to evaluate the clustering performance, in this section, several experiments are conducted by our proposed approach comprehensively comparing with state-of-the-art methods.
The MATLAB codes of our algorithm implementation can be downloaded at \url{http://www.scholat.com/portaldownloadFile.html?fileId=4623}.

\subsection{Datasets}
Four real-world datasets are used to test multi-view data clustering, whose statistics are summarized in Table \ref{DatasetsTab}. The test databases involved are facial, object, digits image and text data. %Some examples of datasets are shown in Fig. \ref{Datasets}.
\begin{table}[htbp]
\centering
\caption{Description of the test datasets.}
\begin{tabular}{cccc}
\toprule
\multicolumn{1}{c}{\multirow{1}{*}{Datasets}} &
\multicolumn{1}{c}{\multirow{1}{*}{No. of samples}} &
\multicolumn{1}{c}{\multirow{1}{*}{No. of views}} &
\multicolumn{1}{c}{\multirow{1}{*}{No. of classes}} \\
\midrule
UCI digits  &2000 &5 & 10\\
Caltech-7     &1474  &6   &7 \\
BBCSport   &544 &2     &5 \\
ORL        &400  &3   &40 \\
\bottomrule
\end{tabular} \label{DatasetsTab}%
\end{table}%
%\begin{figure}[ht]
%\centering
%  \subfigure[]{\includegraphics[width=0.22 \textwidth]{fig_caltech101.pdf}}
%  \subfigure[]{\includegraphics[width=0.18 \textwidth]{fig_ORL.pdf}}\\
%  \subfigure[]{\includegraphics[width=0.35 \textwidth]{fig_UCI.pdf}}
%  \caption{Samples of datasets. (a)Caltech-7; (b)ORL; (c)UCI digits. }\label{Datasets}
%\end{figure}

\textit{UCI digits} is a dataset of handwritten digits of 0 to 9 from UCI machine learning repository \footnote{\url{https://archive.ics.uci.edu/ml/datasets/Multiple+Features}}. It is composed of 2000 data points. In our experiments, 6 published feature sets are utilized to evaluate the clustering performance, including 76 Fourier coefficients of the character shapes (FOU), 216 profile correlations (FAC), 240 pixel averages in $2 \times 3$ windows (Pix), 47 Zernike moment (ZER) and 6 morphological (MOR) features.

\textit{Caltech 101} is an image dataset that consists of 101 categories of images for object
recognition problem. We chose a subset of Caltech 101, called Caltech7, which contains 1474 images of 7 classes, i.e., Face, Motorbikes, Dolla-Bill, Garfield, Snoopy, Stop-Sign and
Windsor-Chair. Six patterns were extracted from all the images, such as Gabor features in dimension of 48 \cite{LadesVorbruggenBuhmannLangeMalsburgWurtzKonen1993}, wavelet moments of dimension 40,  CENTRIST features of dimension 254, histogram of oriented gradients (HoG) features of dimension 1984 \cite{DalalTriggs2005}, GIST features of dimension 512 \cite{OlivaTorralba2001} and local binary patterns (LBP) features of dimension 928 \cite{OjalaPietikainenMaenpaa2002}. %, as shown in Fig. \ref{CalFeas}.

\textit{BBCSport}\footnote{\url{http://mlg.ucd.ie/datasets}} consists of news article data. We select 544 documents from the BBC Sport website corresponding to sports news articles in five topical areas from 2004-2005. It contains 5 class labels, such as athletics, cricket, football, rugby and tennis.

\textit{ORL} face dataset consists of 40 distinct subjects with 10 different images for each. The images are taken at different times with changing lighting conditions, facial expressions and facial details for some subjects. Three types of features, i.e., intensity, LBP features \cite{OjalaPietikainenMaenpaa2002} and Gabor features \cite{LadesVorbruggenBuhmannLangeMalsburgWurtzKonen1993}, are extracted and utilized to test.

%\begin{figure}[ht]
%\centering
%   {\includegraphics[width=0.5 \textwidth]{Cal_feas.pdf}}
%  \caption{Features extracted from Motorbike of Caltech-7. From left to right: Intensity,  ColorMoment, LBP,  HoG, CENTRIST and GIST. }\label{CalFeas}
%\end{figure}

\subsection{Measure metric}
To evaluate all the approaches in terms of clustering,   we here adopt  precision, recall, F-score,
normalized mutual information (NMI), and adjusted rand index (abbreviated to
AR) \cite{HubertArabie1985}, as well as clustering accuracy (ACC). For all these criteria, a higher value means better clustering quality. As each measure penalizes or favors different properties in the clustering, we report results on all the  measures for a comprehensive evaluation.

\subsection{Compared Methods}
Next, we will compare the  proposed method with the following state-of-the-art algorithms, for which there are public code available\footnote{The authors wish to thank these authors for their opening simulation codes.}.
\begin{itemize}
  \item \textit{Single View:} Using the most informative view, i.e., one that achieves the best clustering performance using the graph Laplacian derived from a single view of the data, and performing spectral clustering \cite{CristianiniShawe-TaylorKandola2002} on it.
  \item \textit{Feature Concatenation:} Combining the features of each view one-by-one, and then
conducting spectral clustering, as usual, directly on this concatenated feature representation.
 \item \textit{Kernel Addition:} First building a kernel matrix (affinity matrix) from every feature, and then averaging these matrices to achieve a single kernel matrix input to spectral clustering.
  \item \textit{Centroid based Co-regularized Spectral clustering (CCo-reguSC):} Adopting centroid based co-regularization term to spectral clustering via Gaussian kernel \cite{KumarRaiDaume2011}. The parameter for each view is set to be 0.01 as suggested.
 \item \textit{Pairwise based Co-regularized Spectral clustering (PCo-reguSC):} Adopting pairwise based co-regularization term to spectral clustering via Gaussian kernel \cite{KumarRaiDaume2011}. The parameter for each view is set to be 0.01 as suggested.
  \item \textit{Multi-View NMF (MultiNMF)}\cite{LiuWangGaoHan2013}: In our experiments, we empirically set parameter ($\lambda_v$) to 0.01 for all views and datasets as the authors advised.
  \item \textit{Robust multi-view spectral clustering via Low-Rank and Sparse Decomposition ( LRSD-MSC )}\cite{XiaPanDuYin2014}: This approach recovers a shared low-rank transition probability matrix for multi-view clustering.
  \item \textit{Low-rank tensor constrained multi-view subspace clustering(LT-MSC)}\cite{ZhangFuLiuLiuCao2015}: The method proposes a multi-view clustering by considering the subspace representation matrices of different views as a tensor.
\end{itemize}

In our experiments, \textit{k-means} is utilized at the final step to obtain the clustering results. As \textit{k-means} relies on initialization, we run \textit{k-means} 20 trials and present the means and standard deviations of the performance measures.

\subsection{Performance Evaluation}
In this section, we report the clustering results on the chosen test datasets. In Tables \ref{Tab2}-\ref{Tab5}, the clustering performance by different methods on test datasets are given. The bold numbers highlight the best results. The parameters setting for all the comparing methods is done according to authors' suggestions for their best clustering scores. For the proposed algorithm, we empirically set the parameters and report the results, i.e., $\lambda = 10^{-3}, \alpha = 0.1$ and $\beta = 1.1$. This setting is kept throughout all experiments. As can be seen, our proposed method significantly outperforms other compared ones on all criteria, for all types of data including facial image, object image, digits image and text data. Particularly, for BBCSport, our method outperforms the second best algorithm in terms of ACC/NMI by 19.29\% and 16.23\%, respectively. While for UCI, the leading margins are 10.43\% and 4.76\%, respectively, in terms of ACC/NMI.

LT-MSC achieves the second best result among most cases, especially for the facial image data ORL. This is exactly claimed in \cite{ZhangFuLiuLiuCao2015} and verified in our experiments. While for LRSD-MSC and MultiNMF, they achieved comparable performance. For text data, such as BBCSport, Kernel Addition can produce a better clustering result than other baselines. It is expected that the different multi-view clustering methods may suit varied data. Nevertheless, as it turned out, the proposed method is more suitable and robust for all kinds of multi-view data.

%Next, we further analyze the underlying reasons why the proposed method is superior intuitively. Fig. \ref{Similarity} shows the affinity matrices for UCI digits obtained by several methods. From the figure, we can observe that the subspace within data is well recovered by different methods. Among them, the affinity matrix by our method can better reflect the structure of data, i.e, block-diagonal structure, so as to benefit the subsequent clustering task. On the other hand, the affinity matrix by other methods are somewhat unsatisfactory for clustering.

Furthermore, to show the advantage of combining multi-view features, we choose a part of views of UCI data to form a subset, termed as UCI-2view, which includes 76 Fourier coefficients and 240 pixels. The clustering result is shown in Table \ref{Tab1}. Apparently, the performance degrades when the number of views becomes less, compared to Table \ref{Tab2}. This verifies that the complementary information is indeed beneficial. In other words, multi-view can be employed to comprehensively and accurately describe the data wherever possible \cite{Sun2013}.

\begin{table*}
\caption{Clustering results on UCI database(mean $\pm$ standard deviation). }\label{Tab2}
\begin{center}
\begin{tabular}{|l|c|c|c|c|c|c|}
\hline
\textit{Method} & \textit{ACC} & \textit{F-score} & \textit{Precision} & \textit{Recall} & \textit{NMI} & \textit{AR} \\
\hline\hline
BestView & 0.6956 $\pm$0.0450 &0.5911 $\pm$0.0270 &0.5813 $\pm$0.0268 &0.6014 $\pm$0.0274 &0.6424 $\pm$0.0181 &0.5451 $\pm$0.0300 \\
Feature Concatenation &0.7400 $\pm$0.0004 &0.6470  $\pm$0.0145 & 0.6250$\pm$0.0215 &0.6708$\pm$0.0098 & 0.6973 $\pm$0.0090 & 0.6064 $\pm$0.0167  \\
Kernel Addition &  0.7700 $\pm$0.0006 &0.6954 $\pm$0.0415 & 0.6791$\pm$0.0545 &0.7133$\pm$0.0283 &0.7456 $\pm$0.0193 & 0.6607 $\pm$0.0470 \\
\hline\hline
PCo-reguSC &0.7578$\pm$ 0.0482   &0.6805$\pm$0.0384   &0.6663 $\pm$0.0357  &0.6991$\pm$0.0413   &0.7299  $\pm$0.0336   & 0.6443 $\pm$ 0.0426 \\
CCo-reguSC &0.7667 $\pm$0.0719    &0.7122 $\pm$0.0489   &0.7029$\pm$ 0.0488  & 0.7217$\pm$ 0.0491 &0.7500 $\pm$0.0398   &0.6798$\pm$ 0.0545  \\
MultiNMF  &0.7760 $\pm$ 0.0000   &0.6431  $\pm$0.0000  & 0.6361 $\pm$0.0000  &0.6503$\pm$0.0000   &0.7041  $\pm$0.0000 &0.6031  $\pm$0.0000 \\
LRSD-MSC &0.7700$\pm$ 0.0005   &  0.7095$\pm$ 0.0392 & 0.6915$\pm$0.0444   &0.7286$\pm$0.0352 &0.7581$\pm$0.0244   &0.6764$\pm$0.0440  \\
LT-MSC   & 0.8422$\pm$0.0000   & 0.7828$\pm$0.001  &0.7707$\pm$0.0010  &0.7953$\pm$0.0011  & 0.8217  $\pm$0.0009  &0.7584$\pm$0.0011 \\
SCMV-3DT &\textbf{0.9300 $\pm$0.0000} & \textbf{0.8613 $\pm$0.0004} &  \textbf{0.8591 $\pm$0.0004} & \textbf{0.8635 $\pm$0.0004} & \textbf{0.8608 $\pm$0.0003}  & \textbf{0.8459$\pm$0.0004} \\
\hline
\end{tabular}
\end{center}
\end{table*}

\begin{table*}
\caption{Clustering results on Caltech-7  database(mean $\pm$ standard deviation). }\label{Tab3}
\begin{center}
\begin{tabular}{|l|c|c|c|c|c|c|}
\hline
\textit{Method} & \textit{ACC} & \textit{F-score} & \textit{Precision} & \textit{Recall} & \textit{NMI} & \textit{AR} \\
\hline\hline
BestView & 0.4100$\pm$0.0004& 0.4218$\pm$0.0341 & 0.7353$\pm$0.0406 & 0.2958$\pm$0.0269 &0.4119$\pm$0.0387 &0.2582$\pm$0.0383  \\
Feature Concatenation &0.3800$\pm$0.0001 & 0.3750$\pm$0.0062 &0.6754 $\pm$0.0059 &0.2596 $\pm$0.0063& 0.3410$\pm$0.0045 & 0.2048 $\pm$0.0044 \\
Kernel Addition & 0.3700 $\pm$0.0001 & 0.4163 $\pm$0.0042 & 0.7494$\pm$0.0067 &0.2882$\pm$0.0031 &0.3936 $\pm$0.0214 & 0.2573$\pm$0.0051\\
\hline\hline
PCo-reguSC & 0.4405$\pm$0.0350& 0.4465$\pm$0.0596& 0.7701$\pm$0.1116& 0.3153 $\pm$0.0404& 0.4402$\pm$0.1104&  0.2873$\pm$0.0820\\
CCo-reguSC &0.4222$\pm$0.0334& 0.4456  $\pm$0.0629& 0.7815 $\pm$0.1203& 0.3117 $\pm$0.0423& 0.4564$\pm$0.1251  &0.2894$\pm$0.0856 \\
MultiNMF  & 0.3602$\pm$0.0000 &0.3760$\pm$0.0000 & 0.6486$\pm$0.0000 & 0.2647$\pm$0.0000 & 0.3156 $\pm$0.0000 &0.1965$\pm$0.0000\\
LRSD-MSC &0.4500$\pm$0.0001 &0.4552$\pm$0.0061 &0.7909 $\pm$0.0105 &0.3195$\pm$0.0046 &0.4446 $\pm$0.0052& 0.2998 $\pm$0.0077 \\
LT-MSC   & 0.5665$\pm$0.0001 & 0.5619 $\pm$0.0037& 0.8766$\pm$0.0032 & 0.4135 $\pm$0.0034 & 0.5914 $\pm$0.0073 & 0.4182 $\pm$0.0042\\
SCMV-3DT &\textbf{0.6246$\pm$0.0022} &\textbf{0.6096 $\pm$0.0017} &\textbf{0.8887 $\pm$0.0102} &\textbf{ 0.4640 $\pm$0.0016} &\textbf{0.6031 $\pm$0.0025} &\textbf{0.4693 $\pm$0.0038} \\
\hline
\end{tabular}
\end{center}
\end{table*}

\begin{table*}
\caption{Clustering results on BBCSport database(mean $\pm$ standard deviation). }\label{Tab4}
\begin{center}
\begin{tabular}{|l|c|c|c|c|c|c|}
\hline
\textit{Method} & \textit{ACC} & \textit{F-score} & \textit{Precision} & \textit{Recall} & \textit{NMI} & \textit{AR} \\
\hline\hline
BestView & 0.4300  $\pm$0.0000 & 0.3968  $\pm$0.0017 & 0.2858  $\pm$0.0108 & 0.6549  $\pm$0.0579 & 0.1797   $\pm$0.0126 &0.0973   $\pm$0.0188 \\
Feature Concatenation &0.7200 $\pm$0.0003 &0.6081 $\pm$0.0149 &0.5976 $\pm$0.0385 &0.6234$\pm$0.0408 &0.5524 $\pm$0.0090 &0.4818 $\pm$0.0219 \\
Kernel Addition &0.8200 $\pm$ 0.0001& 0.7496 $\pm$ 0.0092& 0.7725 $\pm$ 0.0171& 0.7285$\pm$ 0.0183& 0.6574 $\pm$ 0.0124 & 0.6741 $\pm$ 0.0116 \\
\hline\hline
PCo-reguSC &0.5335 $\pm$0.0513 & 0.4363  $\pm$0.0212 & 0.3341 $\pm$0.0169 & 0.6343 $\pm$0.0290 &0.2930 $\pm$0.0429 &0.1795 $\pm$0.0316 \\
CCo-reguSC & 0.5140 $\pm$0.0335 &0.4410 $\pm$0.0243 &0.3578 $\pm$0.0307 & 0.6276 $\pm$0.0222 & 0.3283 $\pm$0.0617 &0.2063 $\pm$0.0489 \\
MultiNMF  &0.4467 $\pm$0.0000 & 0.3941  $\pm$0.0000 &0.3246 $\pm$0.0000 &0.5016 $\pm$0.0000 &0.3017 $\pm$0.0000 &0.1471 $\pm$0.0000 \\
LRSD-MSC &0.8215  $\pm$0.0634 & 0.8259  $\pm$0.0468 & 0.8519 $\pm$0.0174 &0.8032 $\pm$0.0739 &0.8013 $\pm$0.0248 &0.7741 $\pm$0.0587 \\
LT-MSC& 0.7169 $\pm$0.0000 & 0.6338  $\pm$0.0000 & 0.5524 $\pm$0.0000 &0.7433 $\pm$0.0000 &0.5565  $\pm$0.0000 &0.4958 $\pm$0.0000 \\
SCMV-3DT & \textbf{0.9800 $\pm$0.0000} & \textbf{0.9505  $\pm$0.0000} & \textbf{0.9594  $\pm$0.0000} & \textbf{0.9418 $\pm$0.0000} & \textbf{0.9298  $\pm$0.0000}  & \textbf{0.9352   $\pm$0.0000} \\
\hline
\end{tabular}
\end{center}
\end{table*}

\begin{table*}
\caption{Clustering results on ORL database(mean $\pm$ standard deviation). }\label{Tab5}
\begin{center}
\begin{tabular}{|l|c|c|c|c|c|c|}
\hline
\textit{Method} & \textit{ACC} & \textit{F-score} & \textit{Precision} & \textit{Recall} & \textit{NMI} & \textit{AR} \\
\hline\hline
BestView  &0.6700 $\pm$0.0000 &0.5787$\pm$0.0554 &0.5154 $\pm$0.0684 &0.6621$\pm$0.0334 & 0.8477 $\pm$0.0182 & 0.5676 $\pm$0.0572 \\
Feature Concatenation &0.6700 $\pm$ 0.0003 & 0.5697  $\pm$ 0.0276 &0.5300 $\pm$ 0.0299 & 0.6160 $\pm$ 0.0264 &0.8329 $\pm$ 0.0116 &0.5590 $\pm$ 0.0284 \\
Kernel Addition &0.6000 $\pm$0.0003 &0.4931 $\pm$0.0265 &0.4324 $\pm$0.0345 &0.5750 $\pm$0.0174 & 0.8062 $\pm$0.0111 &0.4797 $\pm$0.0275 \\
\hline\hline
PCo-reguSC & 0.5827 $\pm$ 0.0231 &0.4609 $\pm$ 0.0171 &0.4021$\pm$ 0.0139 &0.5430 $\pm$ 0.0226 & 0.7859$\pm$ 0.0117 & 0.4465 $\pm$ 0.0175 \\
CCo-reguSC &0.6415 $\pm$0.0324 & 0.5310 $\pm$0.0471 &0.4708 $\pm$0.0427 &0.6103 $\pm$0.0527 &0.8212 $\pm$0.0269 & 0.5187 $\pm$0.0483 \\
MultiNMF   & 0.6825 $\pm$0.0000 &0.5843  $\pm$0.0000 &0.5280$\pm$0.0000 &0.6539 $\pm$0.0000 &0.8393 $\pm$0.0000 &0.5736 $\pm$0.0000 \\
LRSD-MSC  &0.6800 $\pm$0.0485 &0.6047 $\pm$0.0477 &0.5566 $\pm$0.0536 &0.6625 $\pm$0.0391 &0.8515  $\pm$0.0170 &0.5947 $\pm$0.0491 \\
LT-MSC   & 0.7587 $\pm$0.0283 &0.7165 $\pm$0.0232 &0.6540$\pm$0.0263 &0.7926$\pm$0.0232 & \textbf{0.9094 $\pm$0.0094} &0.7093 $\pm$0.0238 \\
SCMV-3DT &\textbf{ 0.7947 $\pm$0.0283} & \textbf{0.7444  $\pm$0.0299} & \textbf{0.6938  $\pm$0.0397} & \textbf{0.8038  $\pm$0.0189} & 0.9088  $\pm$0.0099 & \textbf{0.7381 $\pm$0.0307} \\
\hline
\end{tabular}
\end{center}
\end{table*}

\begin{table*}
\caption{Clustering results on UCI-2view database(mean $\pm$ standard deviation). }\label{Tab1}
\begin{center}
\begin{tabular}{|l|c|c|c|c|c|c|}
\hline
\textit{Method} & \textit{ACC} & \textit{F-score} & \textit{Precision} & \textit{Recall} & \textit{NMI} & \textit{AR} \\
\hline\hline
BestView &0.6800 $\pm$0.0006 &0.5854 $\pm$0.0388 &0.5767 $\pm$0.0380 &0.5944 $\pm$0.0405& 0.6404 $\pm$0.0247 &0.5388 $\pm$0.0431  \\
Feature Concatenation &0.6900 $\pm$0.0006 &0.5906 $\pm$0.0391 &0.5810 $\pm$0.0405 &0.6007 $\pm$0.0380 &0.6415 $\pm$0.0255  &0.5445 $\pm$0.0437\\
Kernel Addition & 0.8300 $\pm$0.0006 &0.7522 $\pm$0.0391 &0.7401 $\pm$0.0520 &0.7651 $\pm$0.0253 & 0.7858 $\pm$0.0212 &0.7241$\pm$0.0441\\
\hline\hline
PCo-reguSC & 0.6905 $\pm$0.0466 &0.5929 $\pm$0.0114 &0.5815 $\pm$0.0124 &0.6054 $\pm$0.0105 &0.6564 $\pm$0.0083 &0.5469 $\pm$0.0128 \\
CCo-reguSC  & 0.8152  $\pm$0.0310 &0.7024 $\pm$0.0429 &0.6957 $\pm$0.0419 &0.7101 $\pm$0.0443 &0.7281  $\pm$0.0318 &0.6691  $\pm$0.0477  \\
MultiNMF  &0.8510 $\pm$0.0000 & 0.7368  $\pm$0.0000 &0.7316 $\pm$0.0000 & 0.7421 $\pm$0.0000 & 0.7650 $\pm$0.0000 &0.7075 $\pm$0.0000 \\
LRSD-MSC  &0.7900 $\pm$0.0006 &0.7054 $\pm$0.0447 & 0.6905 $\pm$0.0531 &0.7213 $\pm$0.0373 & 0.7533 $\pm$0.0298 & 0.6720 $\pm$0.0502 \\
LT-MSC  &0.7680 $\pm$0.0000 & 0.7118 $\pm$0.0000 &0.6970 $\pm$0.0000 & 0.7273 $\pm$0.0000 & 0.7468 $\pm$0.0000 & 0.6792 $\pm$0.0000 \\
SCMV-3DT &\textbf{0.9100  $\pm$0.0000} & \textbf{0.8399  $\pm$0.0002} &\textbf{0.8369 $\pm$0.0003} & \textbf{0.8428 $\pm$0.0001} &  \textbf{0.8414 $\pm$0.0001} & \textbf{0.8221 $\pm$0.0002}\\
\hline
\end{tabular}
\end{center}
\end{table*}

\section{Conclusion}\label{Conclusion}
In this paper, we proposed a novel approach towards low-rank multi-view subspace clustering over third-order tensor data. By using \textit{t-product} based on the circular convolution, the multi-view tensorial data is reconstructed by itself with sparse and low-rank penalty. The proposed method not only takes advantage of the complementary information from multi-view data, but also exploits the multi order correlation consensus. Base on the learned representation, the spectral clustering via Markov chain is applied to final separation subsequently. The extensive experiments, on several multi-view data, are conducted to validate the effectiveness of our approach and demonstrate its superiority against the state-of-the-art methods.

\section*{Acknowledgement}
The authors would like to thank Eric Kernfel for his helpful discussion and Changqing Zhang for his opening code \cite{ZhangFuLiuLiuCao2015}. The Project was supported in part by the Guangdong Natural Science Foundation under Grant (No.2014A030313511), in part by the Scientific Research Foundation for the Returned Overseas Chinese Scholars, State Education Ministry, China.

\end{document}